\documentclass[10pt,twocolumn,twoside]{IEEEtran}
\makeatletter
\long\def\@makecaption#1#2{\ifx\@captype\@IEEEtablestring%
\footnotesize\begin{center}{\normalfont\footnotesize #1}\\
{\normalfont\footnotesize\scshape #2}\end{center}%
\@IEEEtablecaptionsepspace
\else
\@IEEEfigurecaptionsepspace
\setbox\@tempboxa\hbox{\normalfont\footnotesize {#1.}~~ #2}%
\ifdim \wd\@tempboxa >\hsize%
\setbox\@tempboxa\hbox{\normalfont\footnotesize {#1.}~~ }%
\parbox[t]{\hsize}{\normalfont\footnotesize \noindent\unhbox\@tempboxa#2}%
\else
\hbox to\hsize{\normalfont\footnotesize\hfil\box\@tempboxa\hfil}\fi\fi}
\makeatother

\usepackage[pdftex]{graphicx}
\graphicspath{{./fig/}}
\DeclareGraphicsExtensions{.pdf,.jpeg,.png}

\ifCLASSINFOpdf
\else
\fi
\usepackage[cmex10]{amsmath}
\usepackage{amsfonts}
\usepackage{algorithm}
\usepackage{algorithmic}
\usepackage{url}

\begin{document}
%
\title{Clustering on Multi-Layer Graphs via Subspace Analysis on Grassmann Manifolds}
%
%
%

\author{Xiaowen~Dong,
        Pascal~Frossard,
        Pierre~Vandergheynst
        and~Nikolai Nefedov
\thanks{X. Dong, P. Frossard and P. Vandergheynst are with Signal Processing Laboratories (LTS4/LTS2), \'{E}cole Polytechnique F\'{e}d\'{e}rale de Lausanne (EPFL), Lausanne, Switzerland (e-mail: xiaowen.dong@epfl.ch; pascal.frossard@epfl.ch; pierre.vandergheynst@epfl.ch).}
\thanks{N. Nefedov is with Signal and Information Processing Laboratory, Eidgen\"{o}ssische Technische Hochschule Z\"{u}rich (ETH Z\"{u}rich), Zurich, Switzerland (e-mail: nefedov@isi.ee.ethz.ch).}
\thanks{}}

\maketitle

\begin{abstract}
Relationships between entities in datasets are often of multiple nature, like geographical distance, social relationships, or common interests among people in a social network, for example. This information can naturally be modeled by a set of weighted and undirected graphs that form a global multi-layer graph, where the common vertex set represents the entities and the edges on different layers capture the similarities of the entities in term of the different modalities. In this paper, we address the problem of analyzing multi-layer graphs and propose methods for clustering the vertices by efficiently merging the information provided by the multiple modalities. To this end, we propose to combine the characteristics of individual graph layers using tools from subspace analysis on a Grassmann manifold. The resulting combination can then be viewed as a low dimensional representation of the original data which preserves the most important information from diverse relationships between entities. We use this information in new clustering methods and test our algorithm on several synthetic and real world datasets where we demonstrate superior or competitive performances compared to baseline and state-of-the-art techniques. Our generic framework further extends to numerous analysis and learning problems that involve different types of information on graphs.
\end{abstract}


%
\IEEEpeerreviewmaketitle

\section{Introduction}
%
%
%
%
\IEEEPARstart{G}{raphs} are powerful mathematical tools for modeling pairwise relationships among sets of entities; they can be used for various analysis tasks such as classification or clustering. Traditionally, a graph captures a single form of relationships between entities and data are analyzed in light of this one-layer graph. However, numerous emerging applications rely on different forms of information to characterize relationships between entities. Diverse examples include human interactions in a social network or similarities between images or videos in multimedia applications. The multimodal nature of the relationships can naturally be represented by a set of weighted and undirected graphs that share a common set of vertices but with different edge weights depending on the type of information in each graph. This can then be represented by a multi-layer or multi-view graph which gathers all sources of information in a unique representation. Assuming that all the graph layers are informative, they are likely to provide complementary information and thus to offer richer information than any single layer taken in isolation. We thus expect that a proper combination of the information contained in the different layers leads to an improved understanding of the structure of the data and the relationships between entities in the dataset.

\begin{figure}[t]
	\begin{center}
		\begin{tabular}{cc}
			~\includegraphics[width=0.15\textwidth]{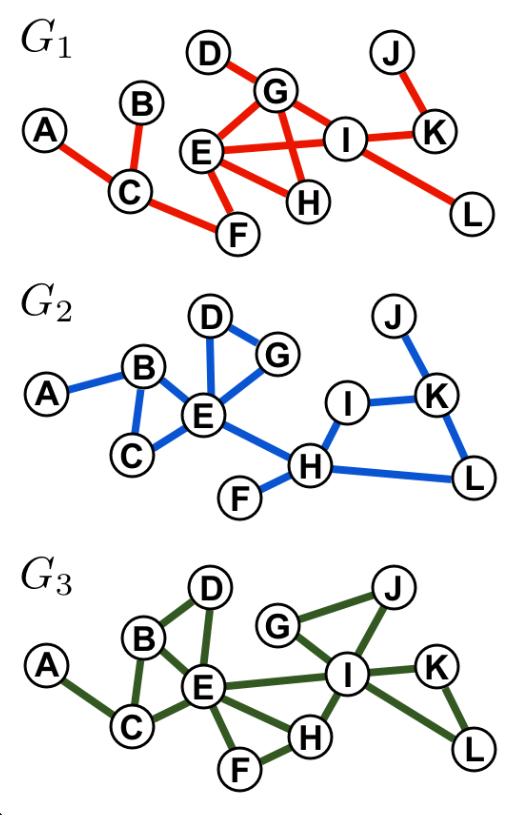}~ & ~\includegraphics[width=0.15\textwidth]{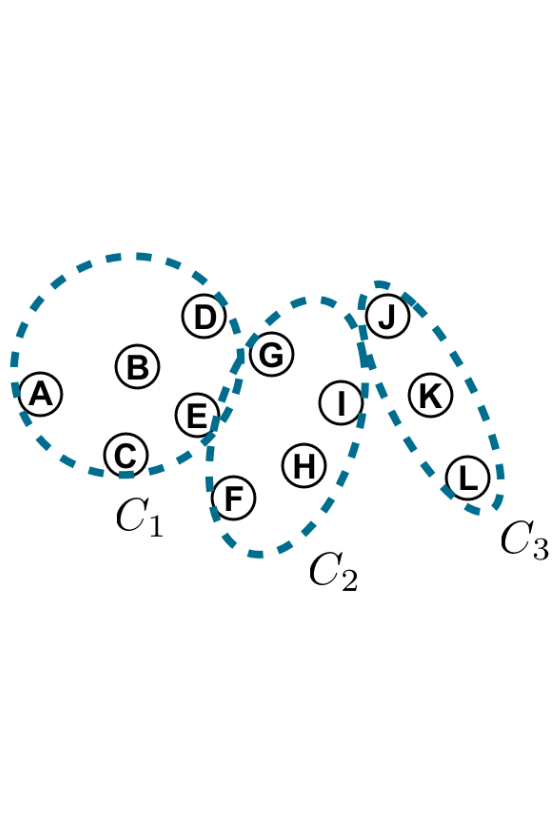}~ \\
			~(a)~ & ~(b)~ \\
		\end{tabular}
	\end{center}
	\caption{(a) An illustration for a three-layer graph $G$, whose three layers $\{G_i\}_{i=1}^3$ share the same set of vertices but with different edges. (b) A potential unified clustering $\{C_k\}_{k=1}^3$ of the vertices based on the information provided by the three layers.}
	\label{fig:multilayer}
\end{figure}

In this paper, we consider a $M$-layer graph $G$ with individual graph layers $G_i=\{V,E_i,\omega_i\}$, $i=1, \ldots, M$, where $V$ represents the common vertex set and $E_i$ represents the edge set in the $i$-th individual graph $G_i$ with associated edge weights $\omega_i$. An example of a three-layer graph is shown in Fig.~\ref{fig:multilayer} (a), where the three graph layers share the same set of 12 vertices but with different edges (we assume unit edge weights for the sake of simplicity). Clearly, different graph layers capture different types of relationships between the vertices, and our objective is to find a method that properly combines the information in these different layers. We first adopt a subspace representation for the information provided by the individual graph layers, which is inspired by the spectral clustering algorithms \cite{Shi00,Ng02,Luxburg07}. We then propose a novel method for combining the multiple subspace representations into one representative subspace. Specifically, we model each graph layer as a subspace on a Grassmann manifold. The problem of combining multiple graph layers is then transformed into the problem of efficiently merging different subspaces on a Grassmann manifold. To this end, we study the distances between the subspaces and develop a new framework to merge the subspaces where the overall distance between the representative subspace and the individual subspaces is minimized. We further show that our framework is well justified by results from statistical learning theory \cite{Hamm09,Gretton05}. The proposed method is a dimensionality reduction algorithm for the original data; it leads to a summarization of the information contained in the multiple graph layers, which reveals the intrinsic relationships between the vertices in the multi-layer graph.

Various learning problems can then be solved using these relationships, such as classification or clustering. Specifically, we focus in this paper on the clustering problem: we want to find a unified clustering of the vertices (as illustrated in Fig.~\ref{fig:multilayer} (b)) by utilizing the representative subspace, such that it is better than clustering achieved on any of the graph layers $G_i$ independently. To address this problem, we first apply our generic framework of subspace analysis on the Grassmann manifold to compute a meaningful summarization (as a representative subspace) of information contained in the individual graph layers. We then implement a spectral clustering algorithm based on the representative subspace. Experiments on synthetic and real world datasets demonstrate the advantages of our approach compared to baseline algorithms, like the summation of individual graphs \cite{Tang09}, as well as state-of-the-art techniques, such as co-regularization \cite{Kumar11a}. Finally, we believe that our framework is beneficial not only to clustering, but also to many other data processing tasks based on multi-layer graphs or multi-view data in general.

This paper is organized as follows. We first review the related work and summarize the contribution of the paper in Section \ref{sec:literature}. In Section \ref{sec:sc}, we describe the subspace representation inspired by spectral clustering, which captures the characteristics of a single graph. In Section \ref{sec:grassmann}, we review the main ingredients of Grassmann manifold theory, and propose a new framework for combining information from multiple graph layers. We then propose our novel algorithm for clustering on multi-layer graphs in Section \ref{sec:clustering}, and compare its performance with other clustering methods on multiple graphs in Section \ref{sec:experiment}. Finally, we conclude in Section \ref{sec:conclusion}.

\section{Related work}
\label{sec:literature}
In this section we review the related work in the literature. First, we describe briefly graph-based clustering algorithms, with a particular focus on the methods that have subspace interpretations. Second, we summarize the previous works built upon subspace analysis and the Grassmann manifold theory. Finally, we report the recent progresses in the field of analysis of multi-layer graphs or multi-view data.

Clustering on graphs has been studied extensively due to its numerous applications in different domains. The works in \cite{Schaeffer07,Fortunato10} have given comprehensive overviews of the advancements in this field over the last few decades. The algorithms that are based on spectral techniques on graphs are of particular interest, typical examples being spectral clustering \cite{Shi00,Ng02,Luxburg07} and modularity maximization via spectral method \cite{Newman06a,Newman06b}. Specifically, these approaches propose to embed the vertices of the original graph into a low dimensional space, usually called the spectral embedding, which consists of the top eigenvectors of a special matrix (graph Laplacian matrix for spectral clustering and modularity matrix for modularity maximization). Due to the special properties of these matrices, clustering in such low dimensional spaces usually becomes trivial. Therefore, the corresponding clustering approaches can be interpreted as transforming the information on the original graph into a meaningful subspace representation. Another example is the Principal Component Analysis (PCA) interpretation on graphs described in \cite{Saerens04}, which links the graph structure to a subspace spanned by the top eigenvectors of the graph Laplacian matrix. These works have inspired us to consider the subspace representation in Section \ref{sec:sc}.

In the past few decades, subspace-based methods have been widely used in classification and clustering problems, most notably in image processing and computer vision. In \cite{Sirovich87,Kirby90}, the authors have discovered that human faces can be characterized by low-dimensional subspaces. In \cite{Turk91}, the authors have proposed to use the so-called ``eigenfaces" for recognition. Inspired by these works, researchers have been particularly interested in data where data points of the same pattern can be represented by a subspace. Due to the growing interests in this field, there is an increasingly large number of works that use tools from the Grassmann manifold theory, which provides a natural tool for subspace analysis. In \cite{Edelman98}, the authors have given a detailed overview of the basics of the Grassmann manifold theory, and developed new optimization techniques on the Grassmann manifold. In \cite{Chikuse03}, the author has presented statistical analysis on the Grassmann manifold. Both works study the distances on the Grassmann manifold. In \cite{Hamm08,Hamm09}, the authors have proposed learning frameworks based on distance analysis and positive semidefinite kernels defined on the Grassmann manifold. Other recent representative works include the studies in \cite{Liu04,Lin06} where the authors have proposed to find optimal subspace representation via optimization on the Grassmann manifold, and the analysis in \cite{Turaga08} where the authors have presented statistical methods on the Stiefel and Grassmann manifolds for applications in vision. Similarly, the work in \cite{Harandi11b} has proposed a novel discriminant analysis framework based on graph embedding for set matching, and the authors in \cite{Wang11} have presented a subspace indexing model on the Grassmann manifold for classification. However, none of the above works considers datasets represented by multi-layer graphs.

At the same time, multi-view data have attracted a large amount of interest in the learning research communities. These data form multi-layer graph representations (or multi-view representations), which generally refer to data that can be analyzed from different viewpoints. In this setting, the key challenge is to combine efficiently the information from multiple graphs (or multiple views) for learning purposes. The existing techniques can be roughly grouped into the following categories. First, the most straightforward way is to form a convex combination of the information from the individual graphs. For example, in \cite{Argyriou05}, the authors have developed a method to learn an optimal convex combination of Laplacian kernels from different graphs. In \cite{Zhou07}, the authors have proposed a Markov mixture model, which corresponds to a convex combination of the normalized adjacency matrices of the individual graphs, for supervised and unsupervised learning. In \cite{Tang12}, the authors have presented several averaging techniques for combining information from the individual graphs for clustering. Second, following the intuitive approaches in the first category, many existing works aim at finding a unified representation of the multiple graphs (or multiple views), but using more sophisticated methods. For instances, the authors in \cite{Tang09,Akata11,Cai11,Dong11a,Eynard12,Liu13} have developed several joint matrix factorization approaches to combine different views of data through a unified optimization framework, where the authors in \cite{Xia10} have proposed to find a unified spectral embedding of the original data by integrating information from different views. Similarly, clustering algorithms based on Canonical Correlation Analysis (CCA) first project the data from different views into a unified low dimensional subspace, and then apply simple algorithms like single linkage or $k$-means to achieve the final clustering \cite{Blaschko08,Chaudhuri09}. Third, unlike the previous methods that try to find a unified representation before applying learning techniques, another strategy in the literature is to integrate the information from individual graphs (views) directly into the optimization problems for the learning purposes. Examples include the co-EM clustering algorithm proposed in \cite{Bickel04}, and the clustering approaches proposed in \cite{Kumar11b,Kumar11a} based on the frameworks of co-training \cite{Blum98} and co-regularization \cite{Sindhwani05}. Fourth, particularly in the analysis of multiple graphs, regularization frameworks on graphs have also been applied. In \cite{Muthukrishnan10}, the authors have presented a regularization framework over edge weights of multiple graphs to compute an improved similarity graph of the vertices (entities). In \cite{Dong11a,Dong12a}, the authors have proposed graph regularization frameworks in both vertex and graph spectral domain to combine individual graph layers. Finally, other representative approaches include the works in \cite{deSa05,Muthukrishnan10} where the authors have defined additional graph representations to incorporate information from the original individual graphs, and the works in \cite{Strehl02,Bruno09,Greene09,Cheng09} where the authors have proposed ensemble clustering approaches by integrating clustering results from individual views. From this perspective, the proposed approach belongs to the second category mentioned above, where we first find a representative subspace for the information provided by the multi-layer graph and then implement the clustering step, or other learning tasks. We believe that this type of approaches is intuitive and easily understandable, yet still flexible and generic enough to be applied to different types of data.

To summarize, the main differences between the related work and the contributions proposed in this paper are the following. First, the research work on Grassmann manifold theory has been mainly focused on subspace analysis. The subspace usually comes directly from the data but are not linked to graph-based learning problems. Our paper makes the explicit link between subspaces and graphs, and presents a fundamental and intuitive way of approaching the learning problems on multi-layer graphs, with help of subspace analysis on the Grassmann manifold. Second, we show the link between the projection distance on the Grassmann manifold \cite{Edelman98,Hamm08} and the empirical estimate of the Hilbert-Schmidt Independence Criterion (HSIC) \cite{Gretton05}. Therefore, together with the results in \cite{Hamm09}, we are able to offer a unified view of concepts from three different perspectives, namely, the projection distance on the Grassmann manifold, the Kullback-Leibler (K-L) divergence \cite{Kullback51} and the HSIC \cite{Gretton05}. This helps to understand better the key concept of distance measure in subspace analysis. Finally, using our novel layer merging framework, we provide a simple yet competitive solution to the problem of clustering on multi-layer graphs. We also discuss the influence of the relationships between the individual graph layers on the performance of the proposed clustering algorithm. We believe that this is helpful towards the design of efficient and adaptive learning algorithms.

\section{Subspace representation for graphs}
\label{sec:sc}
In this section, we describe a subspace representation for the information provided by a single graph. The subspace representation is inspired by spectral clustering, which studies the spectral properties of the graph information for partitioning the vertex set of the graph into several distinct subsets.

Let us consider an weighted and undirected graph $G=(V,E,\omega)$\footnotemark[1], where $V=\{v_i\}_{i=1}^n$ represents the vertex set and $E$ represents the edge set with associated edge weights $\omega$, respectively. Without loss of generality, we assume that the graph is connected. The adjacency matrix $W$ of the graph is a symmetric matrix whose entry $W_{ij}$ represents the edge weight if there is an edge between vertex $v_i$ and $v_j$, or 0 otherwise. The degree of a vertex is defined as the sum of the weights of all the edges incident to it in the graph, and the degree matrix $D$ is defined as the diagonal matrix containing the degrees of each vertex along its diagonal. The normalized graph Laplacian matrix $L$ is then defined as:
\footnotetext[1]{We use the notation $G$ for a single graph exclusively in this section.}
\begin{equation}
L=D^{-\frac{1}{2}}(D-W)D^{-\frac{1}{2}}.
\label{eqn:laplacian}
\end{equation}
The graph Laplacian is of broad interests in the studies of spectral graph theory \cite{Chung97}. Among several variants, we use the normalized graph Laplacian defined in Eq. (\ref{eqn:laplacian}), since its spectrum (i.e., its eigenvalues) always lie between 0 and 2, a property favorable in comparing different graph layers in the following sections. We consider now the problem of clustering the vertices $V=\{v_i\}_{i=1}^n$ of $G$ into $k$ distinct subsets such that the vertices in the same subset are similar, i.e., they are connected by edges of large weights. This problem can be efficiently solved by the spectral clustering algorithms. Specifically, we focus on the algorithm proposed in \cite{Ng02}, which solves the following trace minimization problem:
\begin{equation}
\min_{U \in \mathbb{R}^{n \times k}} tr(U'LU), \quad \mbox{s.t.} \quad U'U=I,
\label{eqn:sc}
\end{equation}
where $n$ is the number of vertices in the graph, $k$ is the target number of clusters, and $(\cdot)'$ denotes the matrix transpose operator. It can be shown by a version of the Rayleigh-Ritz theorem \cite{Luxburg07} that the solution $U$ to the problem of Eq. (\ref{eqn:sc}) contains the first $k$ eigenvectors (which correspond to the $k$ smallest eigenvalues) of $L$ as columns. The clustering of the vertices in $G$ is then achieved by applying the $k$-means algorithm \cite{MacQueen67} to the normalized row vectors of the matrix $U$\footnotemark[2]. As shown in \cite{Luxburg07}, the behavior of spectral clustering can be explained theoretically with analogies to several well-known mathematical problems, such as the normalized graph-cut problem \cite{Shi00}, the random walk process on graphs \cite{Lovasz96}, and problems in perturbation theory \cite{Stewart90,Bhatia97}. This algorithm is summarized in Algorithm~\ref{alg:sc}.
\footnotetext[2]{The necessity for row normalization is discussed in \cite{Luxburg07} and we omit this discussion here. However, the normalization does not change the nature of spectral embedding, hence, it does not affect our derivation later.}

\begin{algorithm}[h]
\caption{Normalized Spectral Clustering \cite{Ng02}}
\begin{algorithmic}[1]

\STATE
\textbf{Input:} \\
$W$: the $n \times n$ weighted adjacency matrix of graph $G$\\
$k$: target number of clusters\\

\STATE
Compute the degree matrix $D$ and the normalized graph Laplacian matrix $L=D^{-\frac{1}{2}}(D-W)D^{-\frac{1}{2}}$.

\STATE
Let $U\in \mathbb{R}^{n \times k}$ be the matrix containing the first $k$ eigenvectors $u_1, \ldots, u_k$ of $L$ (solution of (\ref{eqn:sc})). Normalize each row of $U$ to get $U_\text{norm}$.

\STATE
Let $y_j\in \mathbb{R}^k$ ($j = 1, \ldots, n$) be the $j$-th row of $U_{norm}$.

\STATE
Cluster $y_j$ in $\mathbb{R}^k$ into $k$ clusters $C_1, \ldots, C_k$ using the $k$-means algorithm.

\STATE
\textbf{Output:} \\
$C_1, \ldots, C_k$: the cluster assignment\\

\end{algorithmic}
\label{alg:sc}
\end{algorithm}

\begin{figure}[t]
	\begin{center}
		\begin{tabular}{cc}
			~\includegraphics[width=0.15\textwidth]{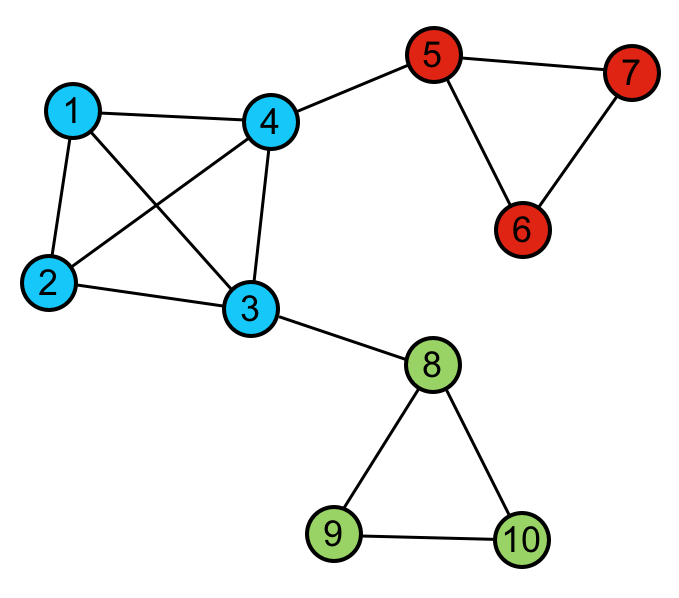}~ & ~\includegraphics[width=0.15\textwidth]{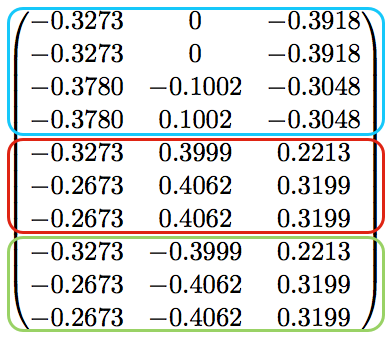}~\\
			~(a)~ \quad & \quad ~(b)~
		\end{tabular}
	\end{center}
	\caption{An illustration of spectral clustering. (a) A graph with three clusters (color-coded) of vertices; (b) Spectral embedding of the vertices computed from the graph Laplacian matrix. The vertices in the same cluster are mapped to coordinates that are close to each other in $\mathbb{R}^3$.}
	\label{fig:toy}
\end{figure}

We provide an illustrative example of the spectral clustering algorithm. Consider a single graph in Fig.~\ref{fig:toy} (a) with ten vertices that belong to three distinct clusters (i.e., $n$=10 and $k$=3). For the sake of simplicity, all the edge weights are set to 1. The low dimensional matrix $U$ that solves the problem of Eq. (\ref{eqn:sc}), which contains $k$ orthonormal eigenvectors of the graph Laplacian $L$ as columns, is shown in Fig.~\ref{fig:toy} (b). The matrix $U$ is usually called the spectral embedding of the vertices, as each row of $U$ can be viewed as the set of coordinates of the corresponding vertex in the $k$-dimensional space. More importantly, due to the properties of the graph Laplacian matrix, such an embedding preserves the connectivity of the vertices in the original graph. In other words, two vertices that are strongly connected in the graph are mapped to two vectors (i.e., rows of $U$) that are close too in the $k$-dimensional space. As a result, a simple $k$-means algorithm can be applied to the normalized row vectors of $U$ to achieve the final clustering of the vertices.

Inspired by the spectral clustering theory, one can define a meaningful subspace representation of the original vertices in a graph by its $k$-dimensional spectral embedding, which is driven by the matrix $U$ built on the first $k$ eigenvectors of the graph Laplacian $L$. Each row being the coordinates of the corresponding vertex in the low dimensional subspace, this representation contains the information on the connectivity of the vertices in the original graph. Such information can be used for finding clusters of the vertices, as shown above, but it is also useful for other analysis tasks on graphs. By adopting this subspace representation that ``summarizes" the graph information, multiple graph layers can naturally be represented by multiple such subspaces (whose geometrical relationships can be quite flexible). The task of multi-layer graph analysis can then be transformed into the problem of effective combination of the multiple subspaces. This is the focus of the next section.

\section{Merging subspaces via analysis on the Grassmann manifold}
\label{sec:grassmann}
We have described above the subspace representation for each graph layer in the multi-layer graph. We discuss now the problem of effectively combining multiple graph layers by merging multiple subspaces. The theory of Grassmann manifold provides a natural framework for such a problem. In this section, we first review the main ingredients of the Grassmann manifold theory, and then move onto our generic framework for merging subspaces.

\subsection{Ingredients of Grassmann manifold theory}
By definition, a Grassmann manifold $\mathcal{G}(k,n)$ is the set of $k$-dimensional linear subspaces in $\mathbb{R}^n$, where each unique subspace is mapped to a unique point on the manifold. As an example, Fig.~\ref{fig:grassmann} shows two 2-dimensional subspaces in $\mathbb{R}^3$ being mapped to two points on $\mathcal{G}(2,3)$. The advantage of using tools from Grassmann manifold theory is thus two-fold: (i) it provides a natural representation for our problem: the subspaces representing the individual graph layers can be considered as different points\footnotemark[3] on the Grassmann manifold; (ii) the analysis on the Grassmann manifold permits to use efficient tools to study the distances between points on the manifold, namely, distances between different subspaces. Such distances play an important role in the problem of merging the information from multiple graph layers. In what follows, we focus on the definition of one particular distance measure between subspaces, which will be used in our framework later on.
\footnotetext[3]{We assume that the Laplacian matrices of any pair of the two layers in the multi-layer graph have different sets of top eigenvectors. In this case, subspace representations for all the layers will be different from each other.}

Mathematically speaking, each point on $\mathcal{G}(k,n)$ can be represented by an orthonormal matrix $Y\in \mathbb{R}^{n \times k}$ whose columns span the corresponding $k$-dimensional subspace in $\mathbb{R}^n$; it is thus denoted as $span(Y)$. For example, the two subspaces shown in Fig.~\ref{fig:grassmann} can be denoted as $span(Y_1)$ and $span(Y_2)$ for two orthonormal matrices $Y_1$ and $Y_2$. The distance between two points on the manifold, or between two subspaces $span(Y_1)$ and $span(Y_2)$, is then defined based on a set of principal angles $\{\theta_i\}_{i=1}^k$ between these subspaces \cite{Golub96}. These principal angles, which measure how the subspaces are geometrically close, are the fundamental measures used to define various distances on the Grassmann manifold, such as the Riemannian (geodesic) distance or the projection distance \cite{Edelman98,Hamm08}. In this paper, we use the projection distance, which is defined as:
\begin{figure}[t]
	\begin{center}
		\begin{tabular}{cc}
			~\includegraphics[width=0.40\textwidth]{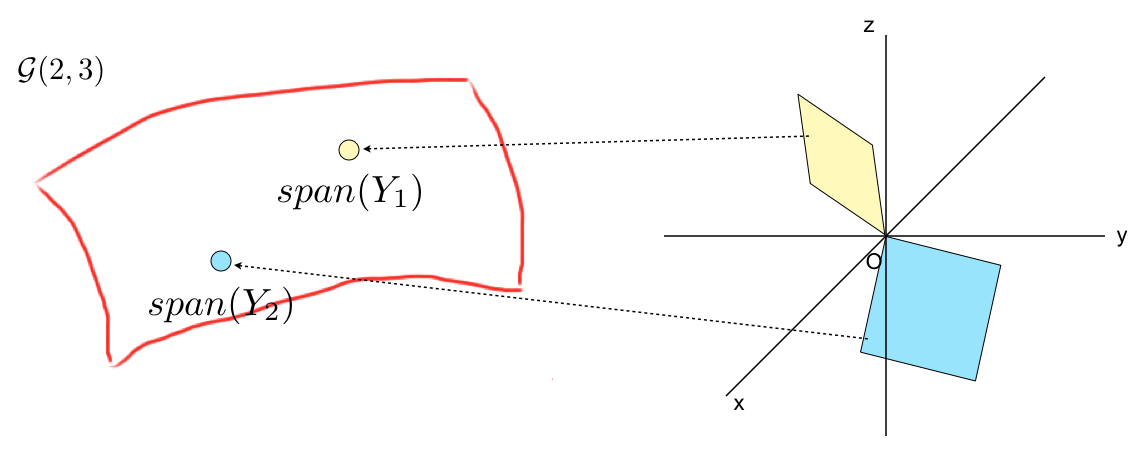}~\\
		\end{tabular}
	\end{center}
	\caption{An example of two 2-dimensional subspaces $span(Y_1)$ and $span(Y_2)$ in $\mathbb{R}^3$, which are mapped to two points on the Grassmann manifold $\mathcal{G}(2,3)$.}
	\label{fig:grassmann}
\end{figure}
\begin{equation}
d_\text{proj}(Y_1,Y_2)=(\sum_{i=1}^k \mbox{sin}^2 \theta_i)^\frac{1}{2},
\end{equation}
where $Y_1$ and $Y_2$ are the orthonormal matrices representing the two subspaces under comparison\footnotemark[4]. The reason for choosing the projection distance is two-fold: (i) the projection distance is defined as the $\ell^2$-norm of the vector of sines of the principal angles. Since it uses all the principal angles, it is therefore an unbiased definition. This is favorable as we do not assume any prior knowledge on the distribution of the data, and all the principal angles are considered to carry meaningful information; (ii) the projection distance can be interpreted using a one-to-one mapping that preserves distinctness: $span(Y) \rightarrow YY' \in \mathbb{R}^{n \times n}$. Note that the squared projection distance can be rewritten as:
\footnotetext[4]{In the special case where $Y_1$ and $Y_2$ represent the same subspace, we have $d_\text{proj}(Y_1,Y_2)=0$.}
\begin{align}
\notag d_\text{proj}^2(Y_1,Y_2)&=\sum_{i=1}^k \mbox{sin}^2 \theta_i \\
\notag &=k-\sum_{i=1}^k \mbox{cos}^2 \theta_i \\
\notag &=k-tr(Y_1{Y_1}'Y_2{Y_2}') \\
\notag &=\frac{1}{2}[2k-2tr(Y_1{Y_1}'Y_2{Y_2}')] \\
\notag &=\frac{1}{2}[tr({Y_1}'Y_1)+tr({Y_2}'Y_2)-2tr(Y_1{Y_1}'Y_2{Y_2}')] \\
&=\frac{1}{2}||Y_1{Y_1}'-Y_2{Y_2}'||_F^2,
\label{eqn:dist}
\end{align}
where the third equality comes from the definition of the principal angles and the fifth equality uses the fact that $Y_1$ and $Y_2$ are orthonormal matrices. It can be seen from Eq. (\ref{eqn:dist}) that the projection distance can be related to the Frobenius norm of the difference between the mappings of the two subspaces $span(Y_1)$ and $span(Y_2)$ in $\mathbb{R}^{n \times n}$. Because the mapping preserves distinctness, it is natural to take the projection distance as a proper distance measure between subspaces. Moreover, the third equality of Eq. (\ref{eqn:dist}) provides an explicit way of computing the projection distance between two subspaces from their matrix representations $Y_1$ and $Y_2$. We are going to use it in developing the generic merging framework in the following section.

To summarize, the Grassmann manifold provides a natural and intuitive representation for subspace-based analysis (as shown in Fig.~\ref{fig:grassmann}). The associated tools, namely the principal angles, permit to define a meaningful distance measure that captures the geometric relationships between the subspaces. Originally defined as a distance measure between two subspaces, the projection distance can be naturally generalized to the analysis of multiple subspaces, as we show in the next section.

\subsection{Generic merging framework}
Equipped with the subspace representation for individual graphs and with a distance measure to compare different subspaces, we are now ready to present our generic framework for merging the information from multiple graph layers. Given a multi-layer graph $G$ with $M$ individual layers $\{G_i\}_{i=1}^M$, we first compute the graph Laplacian matrix $L_i$ for each $G_i$ and then represent each $G_i$ by the spectral embedding matrix $U_i \in \mathbb{R}^{n \times k}$ from the first $k$ eigenvectors of $L_i$, where $n$ is the number of vertices and $k$ is the target number of clusters. Recall that each of the matrices $\{U_i\}_{i=1}^M$ defines a $k$-dimensional subspace in $\mathbb{R}^n$, which can be denoted as $span(U_i)$. The goal is to merge these multiple subspaces in a meaningful and efficient way. To this end, our philosophy is to find a representative subspace $span(U)$ that is close to all the individual subspaces $span(U_i)$, and at the same time the representation $U$ preserves the vertex connectivity in each graph layer. For notational convenience, in the rest of the paper we simply refer to the representations $U$ and $U_i$ as the corresponding subspaces, unless indicated specifically.

The squared projection distance between subspaces defined in Eq. (\ref{eqn:dist}) can be naturally generalized for analysis of multiple subspaces. More specifically, we can define the squared projection distance between the target representative subspace $U$ and the $M$ individual subspaces $\{U_i\}_{i=1}^M$ as the sum of squared projection distances between $U$ and each individual subspace given by $U_i$:
\begin{align}
\notag d_\text{proj}^2(U,\{U_i\}_{i=1}^M)&=\sum_{i=1}^M d_p^2(U,U_i) \\
\notag &=\sum_{i=1}^M [k-tr(UU'U_i{U_i}')] \\
&=kM-\sum_{i=1}^M tr(UU'U_i{U_i}').
\label{eqn:multidist}
\end{align}

\noindent The minimization of the distance measure in Eq. (\ref{eqn:multidist}) enforces the representative subspace $U$ to be close to all the individual subspaces $\{U_i\}_{i=1}^M$ in terms of the projection distance on the Grassmann manifold. At the same time, we want $U$ to preserve the vertex connectivity in each graph layer. This can be achieved by minimizing the Laplacian quadratic form evaluated on the columns of $U$, as also indicated by the objective function in Eq. (\ref{eqn:sc}) for spectral clustering. Therefore, we finally propose to merge multiple subspaces by solving the following optimization problem that integrates Eq. (\ref{eqn:sc}) and Eq. (\ref{eqn:multidist}):
\begin{equation}
\begin{split}
\min_{U \in \mathbb{R}^{n \times k}}\sum_{i=1}^M tr(U'L_iU)+\alpha &[kM-\sum_{i=1}^M tr(UU'U_i{U_i}')], \\
&\mbox{s.t.} \quad U'U=I,
\end{split}
\label{eqn:merge}
\end{equation}
where $L_i$ and $U_i$ are the graph Laplacian and the subspace representation for $G_i$, respectively. The regularization parameter $\alpha$ balances the trade-off between the two terms in the objective function.

The problem of Eq. (\ref{eqn:merge}) can be solved in a similar manner as Eq. (\ref{eqn:sc}). Specifically, by ignoring constant terms and rearranging the trace form in the second term of the objective function, Eq. (\ref{eqn:merge}) can be rewritten as
\begin{equation}
\min_{U \in \mathbb{R}^{n \times k}} tr[U'(\sum_{i=1}^M L_i-\alpha \sum_{i=1}^M U_i{U_i}')U], \quad \mbox{s.t.} \quad U'U=I.
\label{eqn:merge2}
\end{equation}
It is interesting to note that this is the same trace minimization problem as in Eq. (\ref{eqn:sc}), but with a ``modified" Laplacian: 
\begin{equation}
L_\text{mod}=\sum_{i=1}^M L_i-\alpha \sum_{i=1}^M U_i{U_i}'. 
\end{equation}
Therefore, by the Rayleigh-Ritz theorem, the solution to the problem of Eq. (\ref{eqn:merge2}) is given by the first $k$ eigenvectors of the modified Laplacian $L_\text{mod}$, which can be computed using efficient algorithms for eigenvalue problems \cite{Sorensen92,Lehoucq96}.

In the problem of Eq. (\ref{eqn:merge}) we try to find a representative subspace $U$ from the multiple subspaces $\{U_i\}_{i=1}^M$. Such a representation not only preserves the structural information contained in the individual graph layers, which is encouraged by the first term of the objective function in Eq. (\ref{eqn:merge}), but also keeps a minimum distance between itself and the multiple subspaces, which is enforced by the second term. Notice that the minimization of only the first term itself corresponds to simple averaging of the information from different graph layers, which usually leads to suboptimal clustering performance as we shall see in the experimental section. Similarly, imposing only a small projection distance to the individual subspaces $\{U_i\}_{i=1}^M$ does not necessarily guarantee that $U$ is a good solution for merging the subspaces. In fact, for a given $k$-dimensional subspace, there are infinitely many choices for the matrix representation, and not all of them are considered as meaningful summarizations of the information provided by the multiple graph layers. However, under the additional constraint of minimizing the trace of the quadratic term $U'L_iU$ over all the graphs (which is the first term of the objective function in Eq. (\ref{eqn:merge})), the vertex connectivity in the individual graphs tends to be preserved in $U$. In this case, the smaller the projection distance between $U$ and the individual subspaces, the more representative it is for all graph layers.

\subsection{Discussion of the distance function}
Interestingly, the choice of projection distance as a similarity measure between subspaces in the optimization problem of Eq. (\ref{eqn:merge}) can be well justified from information-theoretic and statistical learning points of view. The first justification is from the work of Hamm et al. \cite{Hamm09}, in which the authors have shown that the Kullback-Leibler (K-L) divergence \cite{Kullback51}, which is a well-known similarity measure between two probability distributions in information theory, is closely related to the squared projection distance. More specifically, the work in \cite{Hamm09} suggests that, under certain conditions, we can consider a linear subspace $U_i$ as the ``flattened" limit of a Factor Analyzer distribution $p_i$ \cite{Ghahramani96}:
\begin{equation}
p_i:\mathcal{N}(u_i,C_i), \quad C_i=U_i{U_i}'+\sigma^2I_D,
\end{equation}
where $\mathcal{N}$ stands for the normal distribution, $u_i\in \mathbb{R}^n$ is the mean, $U_i\in \mathbb{R}^{n \times k}$ is a full-rank matrix with $n>k>0$ (which represents the subspace), $\sigma$ is the ambient noise level, and $I_n$ is the identity matrix of dimension $n$. For two subspaces $U_i$ and $U_j$, the symmetrized K-L divergence between the two corresponding distributions $p_i$ and $p_j$ can then be rewritten as:
\begin{equation}
d_\text{KL}(p_1,p_2)=\frac{1}{2\sigma^2(\sigma^2+1)}(2k-2tr(U_i{U_i}'U_j{U_j}')),
\end{equation}
which is of the same form as the squared projection distance when we ignore the constant factor (see Eq. (\ref{eqn:dist})). This shows that, if we take a probabilistic view of the subspace representations $\{U_i\}_{i=1}^M$, then the projection distance between subspaces can be considered consistent with the K-L divergence.

The second justification is from the recently proposed Hilbert-Schmidt Independence Criterion (HSIC) \cite{Gretton05}, which measures the statistical dependence between two random variables. Given $K_{\mathcal{X}_1}, K_{\mathcal{X}_2} \in \mathbb{R}^{n \times n}$ that are the centered Gram matrices of some kernel functions defined over two random variables $\mathcal{X}_1$ and $\mathcal{X}_2$, the empirical estimate of HSIC is given by 
\begin{equation}
d_\text{HSIC}(\mathcal{X}_1,\mathcal{X}_2)=tr(K_{\mathcal{X}_1}K_{\mathcal{X}_2}).
\label{eqn:hsic}
\end{equation}
That is, the larger the $d_\text{HSIC}(\mathcal{X}_1,\mathcal{X}_2)$, the stronger the statistical dependence between $\mathcal{X}_1$ and $\mathcal{X}_2$. In our case, using the idea of spectral embedding, we can consider the rows of the individual subspace representations $U_i$ and $U_j$ as two particular sets of sample points in $\mathbb{R}^k$, which are drawn from two probability distributions governed by the information on vertex connectivity in $G_i$ and $G_j$, respectively. In other words, the sets of rows of $U_i$ and $U_j$ can be seen as realizations of two random variables $\mathcal{X}_i$ and $\mathcal{X}_j$. Therefore, we can define the Gram matrices of linear kernels on $\mathcal{X}_i$ and $\mathcal{X}_j$ as:
\begin{eqnarray}
\notag K_{\mathcal{X}_i}={({U_i}')}'({U_i}')={U_i}{U_i}',\\
K_{\mathcal{X}_j}={({U_j}')}'({U_j}')={U_j}{U_j}'.
\end{eqnarray}
By applying Eq. (\ref{eqn:hsic}), we can see that:
\begin{align}
\notag d_\text{HSIC}(\mathcal{X}_i,\mathcal{X}_j)&=tr(K_{\mathcal{X}_i}K_{\mathcal{X}_j})\\
\notag &=tr({U_i}{U_i}'{U_j}{U_j}')\\
&=k-d_\text{proj}^2(U_i,U_j).
\end{align}
This shows that the projection distance between subspaces $U_i$ and $U_j$ can be interpreted as the negative dependence between $\mathcal{X}_i$ and $\mathcal{X}_j$, which reflect the information provided by the two individual graph layers $G_i$ and $G_j$.

Therefore, from both information-theoretic and statistical learning points of view, the smaller the projection distance between two subspace representations $U_i$ and $U_j$, the more similar the information in the respective graphs that they represent. As a result, the representative subspace (the solution $U$ to the problem of Eq. (\ref{eqn:merge})) can be considered as a subspace representation that ``summarizes" the information from the individual graph layers, and at the same time captures the intrinsic relationships between the vertices in the graph. As one can imagine, such relationships are of crucial importance in our multi-layer graph analysis.

In summary, the concept of treating individual graphs as subspaces, or points on the Grassmann manifold, permits to study the desired merging framework in a unique and principled way. We are able to find a representative subspace for the multi-layer graph of interest, which can be viewed as a dimensionality reduction approach for the original data. We finally remark that the proposed merging framework can be easily extended to take into account the relative importance of each individual graph layer with respect to the specific learning purpose. For instance, when prior knowledge about the importance of the information in the individual graphs is available, we can adapt the value of the regularization parameter $\alpha$ in Eq. (\ref{eqn:merge}) to the different layers such that the representative subspace is closer to the most informative subspace representations.

\section{Clustering on multi-layer graphs}
\label{sec:clustering}
In Section \ref{sec:grassmann}, we introduced a novel framework for merging subspace representations from the individual layers of a multi-layer graph, which leads to a representative subspace that captures the intrinsic relationships between the vertices of the graph. This representative subspace provides a low dimensional form that can be used in several applications involving multi-layer graph analysis. In particular, we study now one such application, namely the problem of clustering vertices in a multi-layer graph. We further analyze the behavior of the proposed clustering algorithm with respect to the properties of the individual graph layers (subspaces).

\subsection{Clustering algorithm}
As we have already seen in Section \ref{sec:sc}, the success of the spectral clustering algorithm relies on the transformation of the information contained in the graph structure into a spectral embedding computed from the graph Laplacian matrix, where each row of the embedding matrix (after normalization) is treated as the coordinates of the corresponding vertex in a low dimensional subspace. In our problem of clustering on a multi-layer graph, the setting is slightly different, since we aim at finding a unified clustering of the vertices that takes into account information contained in all the individual layers of the multi-layer graph. However, the merging framework proposed in the previous section can naturally be applied in this context. In fact, it leads to a natural solution to the clustering problem on multi-layer graphs. In more details, similarly to the spectral embedding matrix in the spectral cluttering algorithm, which is a subspace representation for one individual graph, our merging framework provides a representative subspace that contains the information from the multiple graph layers. Using this representation, we can then follow the same steps of spectral clustering to achieve the final clustering of the vertices with a $k$-means algorithm. The proposed clustering algorithm is summarized in Algorithm~\ref{alg:merge}.

\begin{algorithm}[h]
\caption{Spectral Clustering on Multi-Layer graphs (\textbf{SC-ML})}

\begin{algorithmic}[1]

\STATE
\textbf{Input:} \\
$\{W_i\}_{i=1}^M$: $n \times n$ weighted adjacency matrices of individual graph layers $\{G_i\}_{i=1}^M$\\
$k$: target number of clusters\\
$\alpha$: regularization parameter\\

\STATE
Compute the normalized Laplacian matrix $L_i$ and the subspace representation $U_i$ for each $G_i$.

\STATE
Compute the modified Laplacian matrix $L_{\text{mod}}=\sum_{i=1}^M L_i-\alpha \sum_{i=1}^M {U_i}{U_i}'$.

\STATE
Compute $U\in \mathbb{R}^{n \times k}$ that is the matrix containing the first $k$ eigenvectors $u_1, \ldots, u_k$ of $L_{\text{mod}}$. Normalize each row of $U$ to get $U_{\text{norm}}$.

\STATE
Let $y_j\in \mathbb{R}^k$ ($j = 1, \ldots, n$) be the $j$-th row of $U_{\text{norm}}$.

\STATE
Cluster $y_j$ in $\mathbb{R}^k$ into $C_1, \ldots, C_k$ using the $k$-means algorithm.

\STATE
\textbf{Output:} \\
$C_1, \ldots, C_k$: The cluster assignment\\

\end{algorithmic}
\label{alg:merge}
\end{algorithm}

It is clear that Algorithm~\ref{alg:merge} is a direct generalization of Algorithm~\ref{alg:sc} in the case of multi-layer graphs. The main ingredient of our clustering algorithm is the merging framework proposed in Section \ref{sec:grassmann}, in which information from individual graph layers is summarized, prior to the actual clustering process (i.e., the $k$-means step) is implemented. This provides an example that illustrates how our generic merging framework can be applied to specific learning tasks on multi-layer graphs.

\subsection{Analysis of the proposed algorithm}
We now analyze the behavior of the proposed clustering algorithm under different conditions. Specifically, we first outline the link between subspace distance and clustering quality, and then compare the clustering performances in two scenarios where the relationships between the individual subspaces $\{U_i\}_{i=1}^M$ are different.

As we have seen in Section \ref{sec:grassmann}, the rows of the subspace representations $\{U_i\}_{i=1}^M$ can be viewed as realizations of random variables $\{{\mathcal{X}_i}\}_{i=1}^M$ governed by the graph information. At the same time, spectral clustering directly utilizes $U_i$ for the purpose of clustering. Therefore, $\{{\mathcal{X}_i}\}_{i=1}^M$ can be considered as random variables that control the cluster assignment of the vertices. In fact, it has been shown in \cite{Luxburg07} that the matrix $U_i$ is closely related to the matrix that contains the cluster indicator vectors as columns. Since the projection distance can be understood as the negative statistical dependence between such random variables, the minimization of the projection distance in Eq. (\ref{eqn:merge}) is equivalent to the maximization of the dependence between the random variable from the representative subspace $U$ and the ones from the individual subspaces $\{U_i\}_{i=1}^M$. The optimization in Eq. (\ref{eqn:merge}) can then be seen as a solution that tends to produce a clustering with the representative subspace that is consistent with those computed from the individual subspace representations.

\begin{figure}[t]
	\begin{center}
		\begin{tabular}{cc}
			~\includegraphics[width=0.12\textwidth]{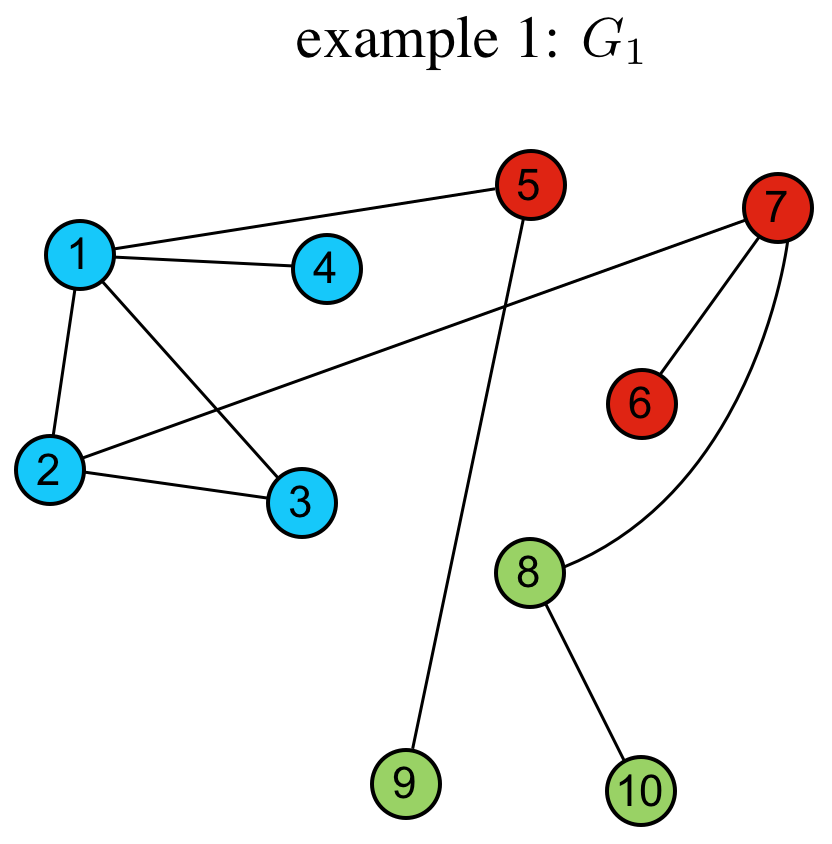}~ ~\includegraphics[width=0.12\textwidth]{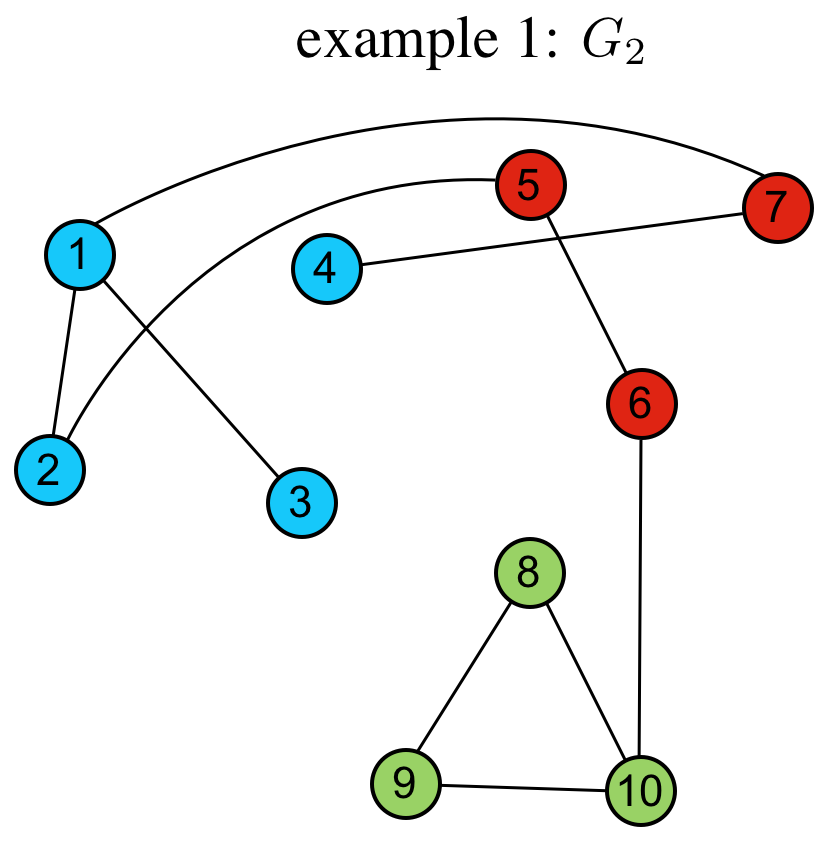}~  ~\includegraphics[width=0.12\textwidth]{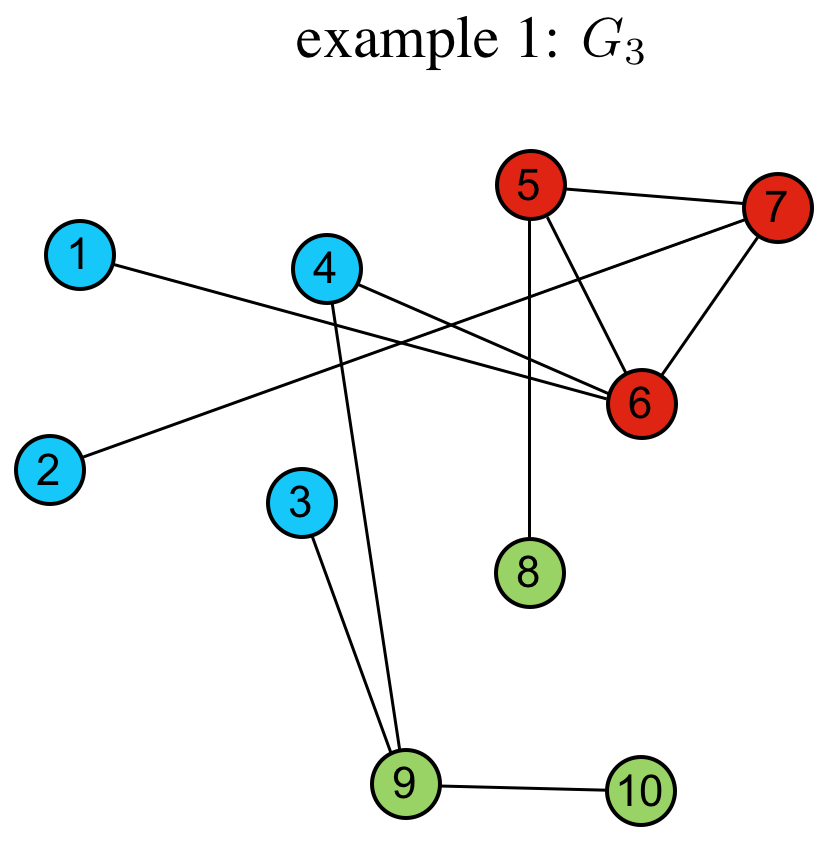}~ \\
		\end{tabular}
	\end{center}
	\caption{A 3-layer graph with unit edge weights for toy example 1. The colors indicate the groundtruth clusters.}
	\label{fig:toy1}
\end{figure}

\begin{table}[t]
	\caption{Analysis of toy example 1.}
	\begin{center}
		\begin{tabular}{cc}
			~\includegraphics[width=0.40\textwidth]{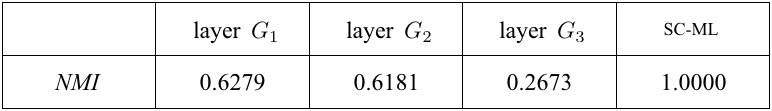}~ \\
			~(a) clustering performances for toy example 1~ \vspace{0.2cm}\\
			~\includegraphics[width=0.40\textwidth]{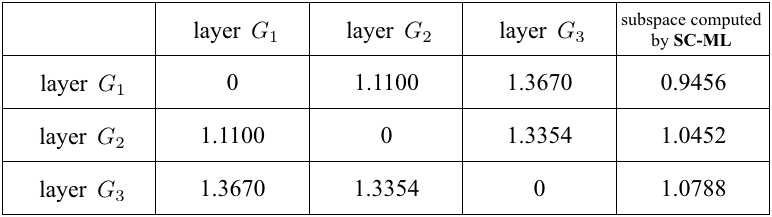}~ \\
			~(b) subspace distances for toy example 1~\\
		\end{tabular}
	\end{center}
	\label{tab:toy1}
\end{table}

\begin{figure}[t]
	\begin{center}
		\begin{tabular}{cc}
			~\includegraphics[width=0.12\textwidth]{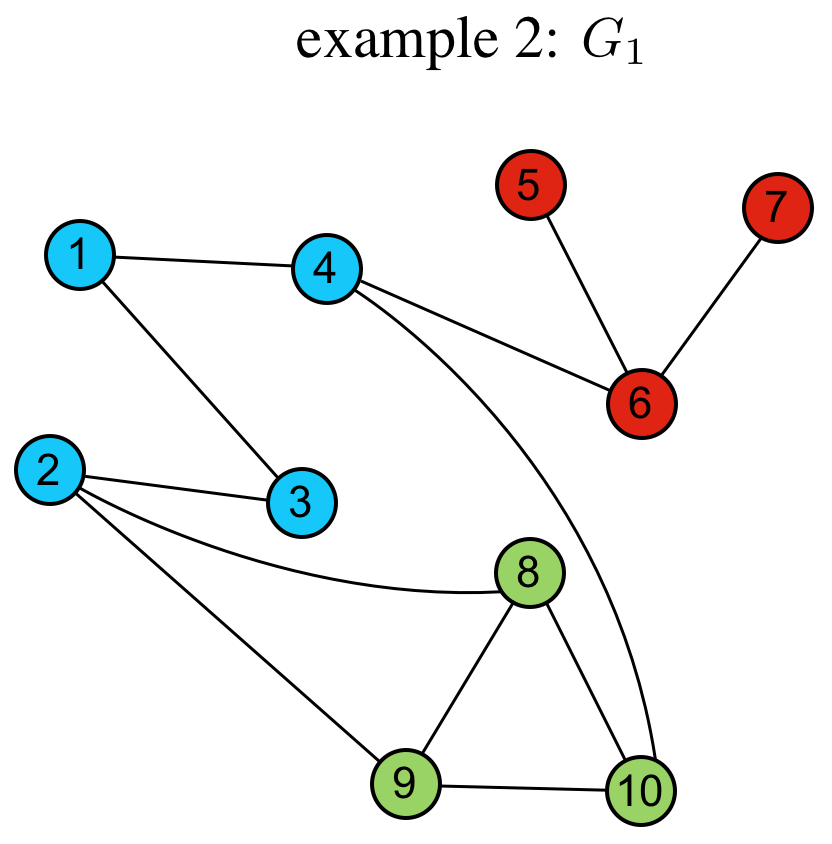}~ ~\includegraphics[width=0.12\textwidth]{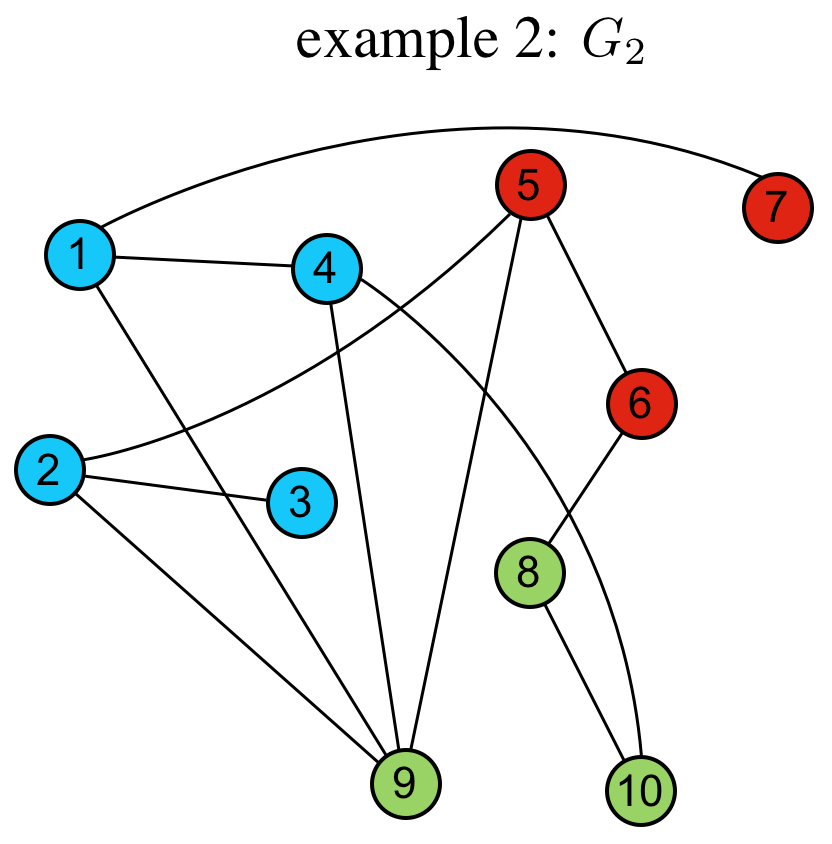}~  ~\includegraphics[width=0.12\textwidth]{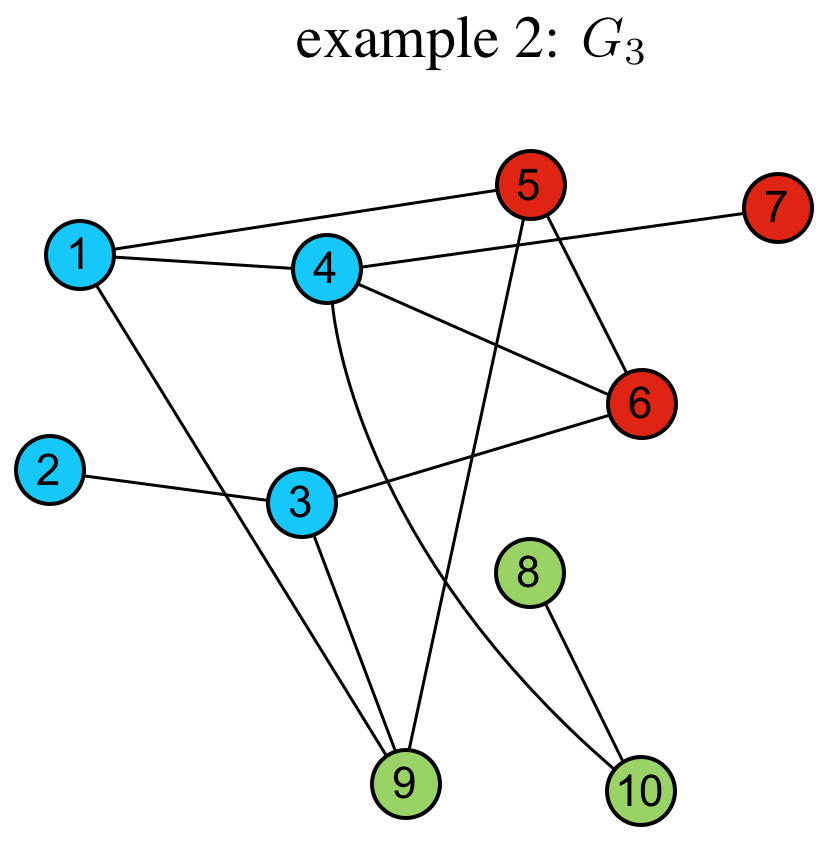}~ \\
		\end{tabular}
	\end{center}
	\caption{A 3-layer graph with unit edge weights for toy example 2. The colors indicate the groundtruth clusters.}
	\label{fig:toy2}
\end{figure}

We now discuss how the relationships between the individual subspaces possibly affect the performance of our clustering algorithm \textbf{SC-ML}. Intuitively, since the second term of the objective function in Eq. (\ref{eqn:merge}) represents the distance between the representative subspace $U$ and all the individual subspaces $\{U_i\}_{i=1}^M$, it tends to drive the solution towards those subspaces that themselves are close to each other on the Grassmann manifold. To show it more clearly, let us consider two toy examples. The first example is illustrated in Fig.~\ref{fig:toy1}, where we have a 3-layer graph with the individual layers $G_1$, $G_2$ and $G_3$ sharing the same set of vertices. For the sake of simplicity, all the edge weights are set to one. In addition, three groundtruth clusters are indicated by the colors of the vertices. Table~\ref{tab:toy1} (a) shows the performances of Algorithm~\ref{alg:sc} with individual layers as well as Algorithm~\ref{alg:merge}\footnotemark[5] for the multi-layer graph, in terms of \textit{Normalized Mutual Information (NMI)} \cite{Manning08} with respect to the groundtruth clusters. Table~\ref{tab:toy1} (b) shows the projection distances between various pairs of subspaces. It is clear that the layers $G_1$ and $G_2$ produce better clustering quality, and that the distance between the corresponding subspaces is smaller. However, the vertex connectivity in layer $G_3$ is not very consistent with the groundtruth clusters and the corresponding subspace is further away from the ones from $G_1$ and $G_2$. In this case, the solution found by \textbf{SC-ML} is enforced to be close to the consistent subspaces from $G_1$ and $G_2$, hence provides satisfactory clustering results ($\textit{NMI}=1$ represents perfect recovery of groundtruth clusters).
Let us now consider a second toy example, as illustrated in Fig.~\ref{fig:toy2}. In this example we have two layers $G_2$ and $G_3$ with relatively low quality information with respect to the groundtruth clustering of the vertices. As we see in Table~\ref{tab:toy2} (b), their corresponding subspaces are close to each other on the Grassmann manifold. The most informative layer $G_1$, however, represents a subspace that is quite far away from the ones from $G_2$ and $G_3$. At the same time, we see in Table~\ref{tab:toy2} (a) that the clustering results are better for the first layer than for the other two less informative layers. If the quality of the information in the different layers is not considered in computing the representative subspace, \textbf{SC-ML} enforces the solution to be closer to two layers of relatively lower quality, which results in unsatisfactory clustering performance in this case.
\footnotetext[5]{We choose the value of the regularization parameter that leads to the best possible clustering performance. More discussions about the choices of this parameter are presented in Section \ref{sec:experiment}.}

\begin{table}[t]
	\caption{Analysis of toy example 2.}
	\begin{center}
		\begin{tabular}{cc}
			~\includegraphics[width=0.40\textwidth]{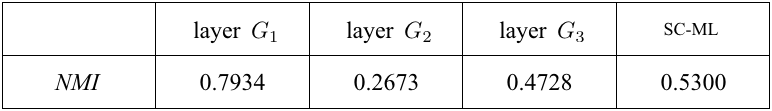}~ \\
			~(a) clustering performances for toy example 2~ \vspace{0.2cm}\\
			~\includegraphics[width=0.40\textwidth]{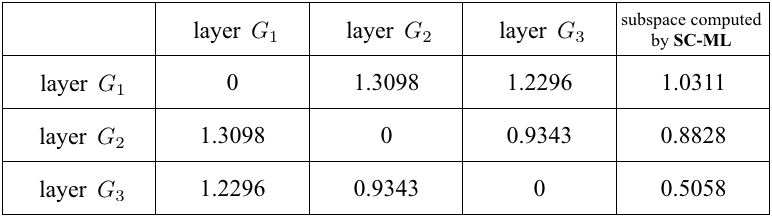}~ \\
			~(b) subspace distances for toy example 2~\\
		\end{tabular}
	\end{center}
	\label{tab:toy2}
\end{table}

The analysis above implies that the proposed clustering algorithm works well under the following assumptions: (i) the majority of the individual subspaces are relatively informative, namely, they are helpful for recovering the groundtruth clustering, and (ii) they are reasonably close to each other on the Grassmann manifold, namely, they provide \textit{complementary} but not \textit{contradictory} information. These are the assumptions made in the present work. As we shall see in the next section, these assumptions seem to be appropriate and realistic in real world datasets. If it is not the case, one may assume that a preprocessing step cleans the datasets, or at least provides information about the reliability of the information in the different graph layers.

\section{Experimental results}
\label{sec:experiment}
In this section, we evaluate the performance of the \textbf{SC-ML} algorithm presented in Section \ref{sec:clustering} on several synthetic and real world datasets. We first describe the datasets that we use for the evaluation, and then explain the various clustering algorithms that we adopt in the performance comparisons. We finally present the results in terms of three evaluation criteria as well as some discussions.

\subsection{Datasets}
We adopt one synthetic and two real world datasets with multi-layer graph representation for the evaluation of the clustering algorithms. We give a brief overview of the datasets as follows.

\begin{figure}[t]
	\begin{center}
		\begin{tabular}{cc}
			~\includegraphics[width=0.40\textwidth]{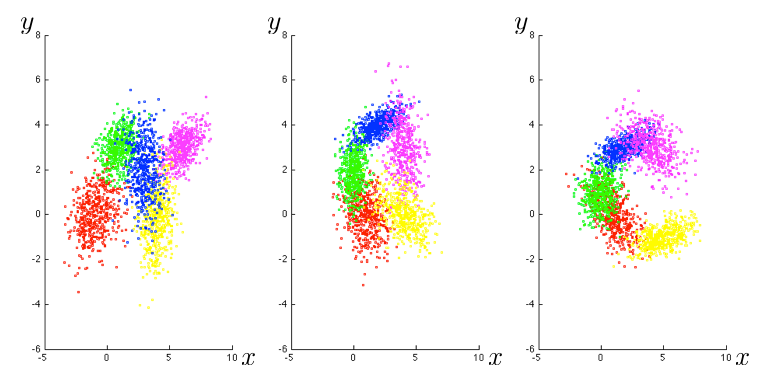}~\\
		\end{tabular}
	\end{center}
	\caption{Three five-class point clouds in $\mathbb{R}^2$ forming English letters ``N", ``R" and ``C".}
	\label{fig:nrc}
\end{figure}

The first dataset that we use is a synthetic dataset, where we have three point clouds in $\mathbb{R}^2$ forming the English letters ``N", ``R" and ``C" (shown in Fig.~\ref{fig:nrc}). Each point cloud is generated from a five-component Gaussian mixture model with different values for the mean and variance of the Gaussian distributions, where each component represents a class of 500 points with specific color. A 5-nearest neighbor graph is then constructed for each point cloud by assigning the weight of the edges connecting two vertices (points) as the reciprocal of the Euclidean distance between them. This gives us a 3-layer graph of 2500 vertices, where each graph layer is from a point cloud forming a particular letter. The goal with this dataset is to recover the five clusters (indicated by five colors) of the 2500 vertices using the three graph layers constructed from the three point clouds.

The second dataset contains data collected during the Lausanne Data Collection Campaign \cite{Kiukkonen10} by the Nokia Research Center (NRC) in Lausanne. This dataset contains the mobile phone data of 136 users living and working in the Lake L\'{e}man region in Switzerland, recorded over a one-year period. Considering the users as vertices in the graph, we construct three graphs by measuring the proximities between these users in terms of GPS locations, Bluetooth scanning activities and phone communication. More specifically, for GPS locations and bluetooth scans, we measure how many times two users are sufficiently close geographically (within a distance of roughly 1 km), and how many times two users' devices have detected the same bluetooth devices, respectively, within 30-minute time windows. Aggregating these results for a one-year period leads to two weighted adjacency matrices that represent the physical proximities of the users measured with different modalities. In addition, an adjacency matrix for phone communication is generated by assigning edge weights depending on the number of calls between any pair of two users. These three adjacency matrices form a 3-layer graph of 136 vertices, where the goal is to recover the eight groundtruth clusters that have been constructed from the users' email affiliations.

The third dataset is a subset of the Cora bibliographic dataset\footnotemark[6]. This dataset contains 292 research papers from three different fields, namely, natural language processing, data mining and robotics. Considering papers as vertices in the graph, we construct the first two graphs by measuring the similarities among the title and the abstract of these papers. More clearly, for both title and abstract, we represent each paper by a vector of non-trivial words using the \textit{Term Frequency-Inverse Document Frequency (TF-IDF)} weighting scheme, and compute the cosine similarities between every pair of vectors as the edge weights in the graphs. Moreover, we add a third graph which reflects the citation relationships among the papers, namely, we assign an edge with unit weight between papers $A$ and $B$ if $A$ has cited or been cited by $B$. This results in a 3-layer graph of 292 vertices, and the goal in this dataset is to recover the three clusters corresponding to the different fields the papers belong to.
\footnotetext[6]{Available online at ``\url{http://people.cs.umass.edu/~mccallum/data.html}" under category ``Cora Research Paper Classification".}

To visualize the graphs in the three datasets, the spy plot of the adjacency matrices of the graphs are shown in Fig.~\ref{fig:spy} (a), (b) and (c) for the synthetic, NRC and Cora dataset, respectively, where the orderings of the vertices are made consistent with the groundtruth clusters\footnotemark[7]. A spy plot is a global view of a matrix where every non-zero entry in the matrix is represented by a blue dot (without taking into account the value of the entry). As shown in these figures, we see clearly the clusters in the synthetic and Cora datasets, while the clusters in the NRC dataset are not very clear. The reason for this is that, in the NRC dataset, the email affiliations used to create the groundtruth clusters only provides approximative information.
\footnotetext[7]{The adjacency matrix for GPS proximity in the NRC dataset is thresholded for better illustration.}

\subsection{Clustering algorithms}
We now explain briefly the clustering algorithms in our comparative performance analysis along with some implementation details. We adopt three baseline algorithms as well as a state-of-the-art technique, namely the co-regularization approach introduced in \cite{Kumar11a}. As we shall see, there is an interesting connection between this approach and the proposed algorithm. First of all, we describe some implementation details of the proposed \textbf{SC-ML} algorithm and the co-regularization approach in \cite{Kumar11a}:

\begin{figure*}[t]
	\begin{center}
		\begin{tabular}{cc}
			~\includegraphics[width=0.60\textwidth]{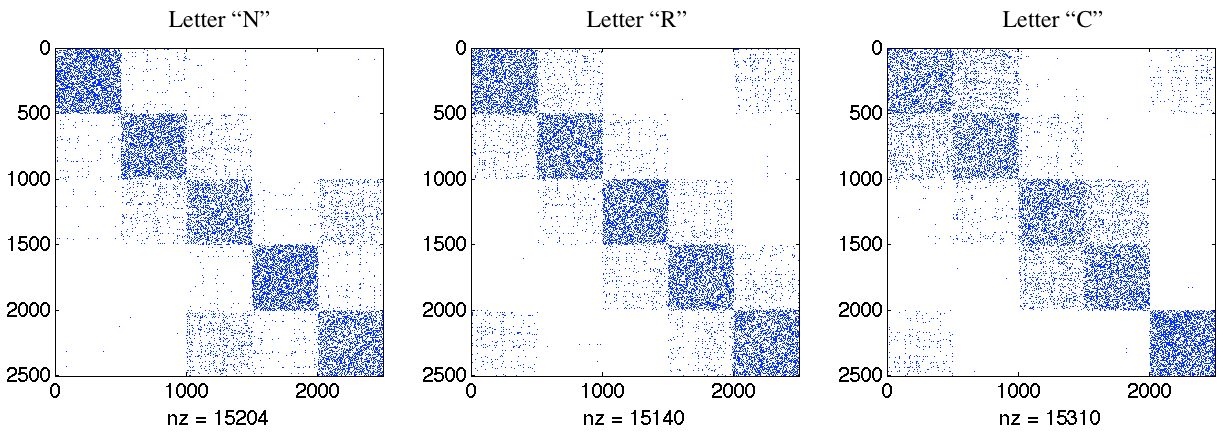}~ \\
			~(a)~ \\
			~\includegraphics[width=0.60\textwidth]{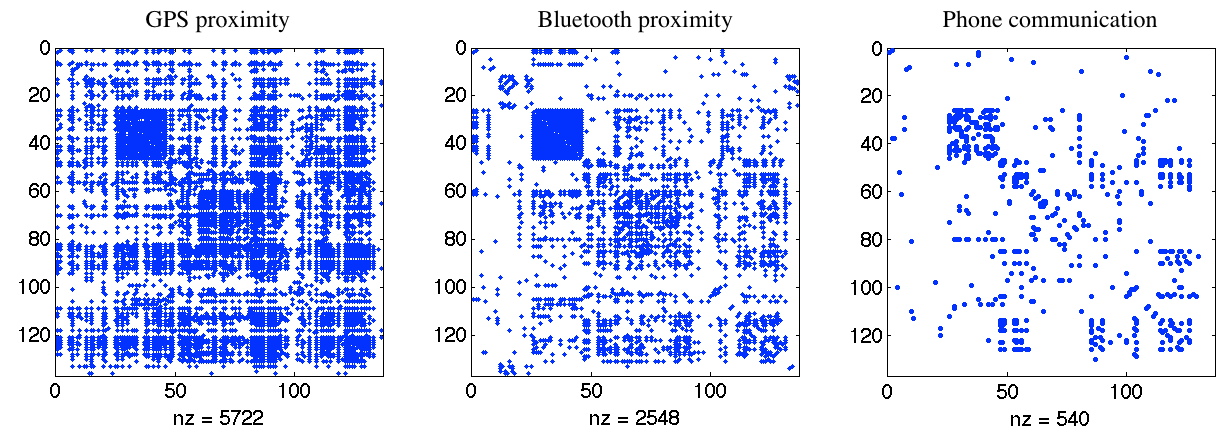}~ \\
			~(b)~ \\
			~\includegraphics[width=0.60\textwidth]{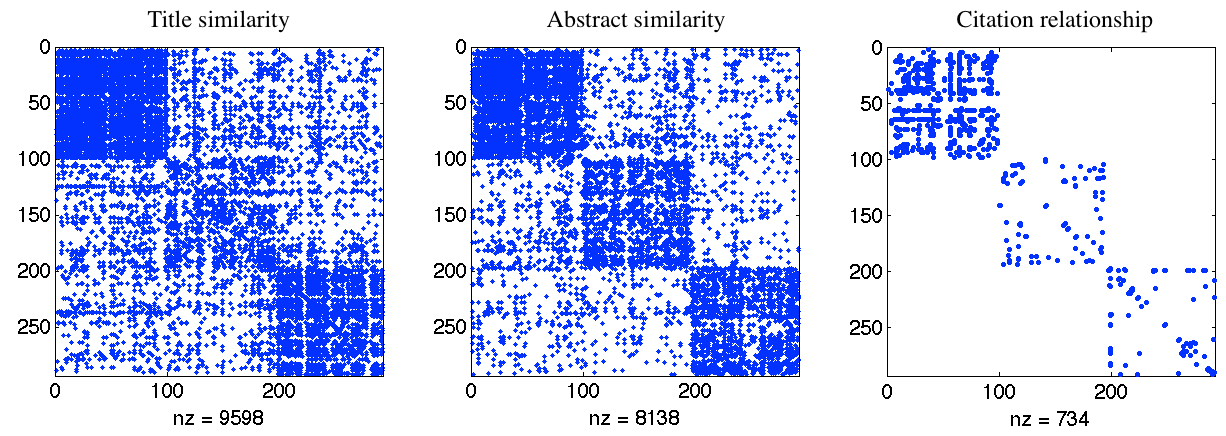}~ \\
			~(c)~ \\
		\end{tabular}
	\end{center}
	\vspace{-0.2cm}
	\caption{Spy plots of three adjacency matrices in (a) the synthetic dataset, (b) the NRC dataset, and (c) the Cora dataset.}
	\label{fig:spy}
\end{figure*}

\begin{itemize}
\item \textbf{SC-ML}: Spectral Clustering on Multi-Layer graphs, as presented in Section \ref{sec:clustering}. The implementation of \textbf{SC-ML} is pretty straightforward, and the only parameter to choose is the regularization parameter $\alpha$ in Eq. (\ref{eqn:merge}). In our experiments, we choose the value of $\alpha$ through multiple empirical trials and report the best clustering performance. Specifically, we choose $\alpha$ to be 0.64 for the synthetic dataset and 0.44 for both real world datasets. We will discuss the choice of this parameter later in this section.

\item \textbf{SC-CoR}: Spectral Clustering with Co-Regularization proposed in \cite{Kumar11a}. We follow the same practice as in \cite{Kumar11a} to choose the most informative graph layer to initialize the alternating optimization scheme in \textbf{SC-CoR}. The stopping criteria for the optimization process is chosen such that the optimization stops when changes in the objective function are smaller than $10^{-5}$. Similarly, we choose the value of the regularization parameter $\alpha$ in \textbf{SC-CoR} through multiple empirical trials and report the best clustering performance. As in \cite{Kumar11a}, the parameter $\alpha$ is fixed in the optimization steps for all graph layers.
\end{itemize}

\begin{table*}[h!tbp]
	\caption{Performance comparison of different clustering algorithms on (a) the synthetic dateset, (b) the NRC dataset, and (c) the Cora dataset.}
	\begin{center}
		\begin{tabular}{cc}
			~\includegraphics[width=0.5\textwidth]{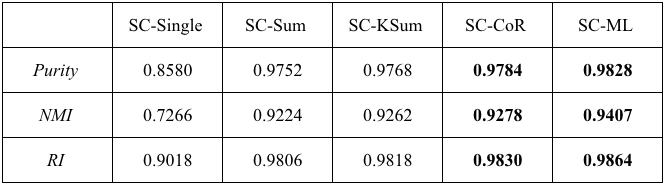}~ \\
			~(a)~ \\
			~\includegraphics[width=0.5\textwidth]{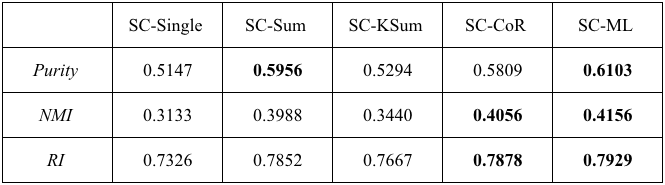}~ \\
			~(b)~ \\
			~\includegraphics[width=0.5\textwidth]{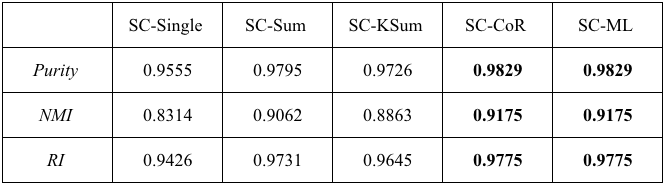}~ \\
			~(c)~ \\
		\end{tabular}
	\end{center}
	\label{tab:results}
\end{table*}

Next, we introduce three baseline comparative algorithms that work as follows:
\begin{itemize}
\item \textbf{SC-Single}: Spectral Clustering (Algorithm~\ref{alg:sc}) applied on a single graph layer, where the graph is chosen to be the one that leads to the best clustering results.
\item \textbf{SC-Sum}: Spectral clustering applied on a global matrix $W$ that is the summation of the normalized adjacency matrices of the individual layers:
\begin{equation}
W=\sum_{i=1}^M D_i^{-\frac{1}{2}}W_i D_i^{-\frac{1}{2}}.
\end{equation}
\item \textbf{SC-KSum}: Spectral clustering applied on the summation $K$ of the spectral kernels \cite{Tang09} of the adjacency matrices: 
\begin{equation}
K= \sum_{i=1}^M K_i \quad \mbox{with} \quad K_i=\sum_{m=1}^{d}u_{im}{u_{im}}',
\end{equation}
where $n$ is the number of vertices, $d\ll n$ is the number of eigenvectors used in the definition of the spectral kernels $K_i$, and $u_{im}$ represents the $m$-th eigenvector of the Laplacian $L_i$ for graph $G_i$. To make it more comparable with spectral clustering, we choose $d$ to be the target number of clusters in our experiments.
\end{itemize}

\subsection{Results and discussions}
We evaluate the performance of the different clustering algorithms with three different criteria, namely \textit{Purity}, \textit{Normalized Mutual Information (NMI)} and \textit{Rand Index (RI)} \cite{Manning08}. The results are summarized in Table~\ref{tab:results} (a), (b) and (c) for the synthetic, NRC and Cora dataset, respectively. For each scenario, the best two results are highlighted in bold fonts. First, as expected, we see that the clustering performances for the synthetic and Cora datasets are higher than that for the NRC dataset, which indicates that the latter one is indeed more challenging due to the approximative groundtruth information. Second, it is clear that \textbf{SC-ML} and \textbf{SC-CoR} generally outperform the baseline approaches for the three datasets. More specifically, although both \textbf{SC-Sum} and \textbf{SC-KSum} indeed improve the clustering quality compared to clustering with individual graph layers, they only provide limited improvement, and the potential drawback for both of the summation methods is that they can be considered as similar to building a simple average graph for representing the different layers of information. Therefore, depending on data characteristics in specific datasets, this might smooth out the particular information provided by individual layers, and thus penalize the clustering performance. In comparison, \textbf{SC-ML} and \textbf{SC-CoR} always achieve significant improvements in the clustering quality compared to clustering using individual graph layers.

\begin{figure*}[t]
	\begin{center}
		\begin{tabular}{cc}
			~\includegraphics[width=0.66\textwidth]{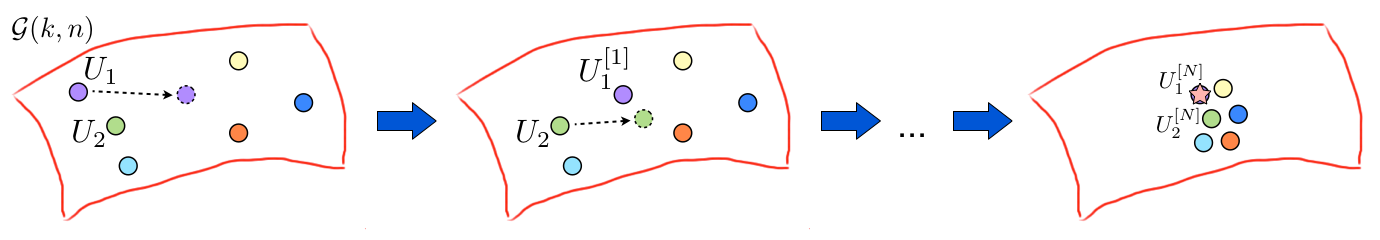}~ \quad & \quad ~\includegraphics[width=0.18\textwidth]{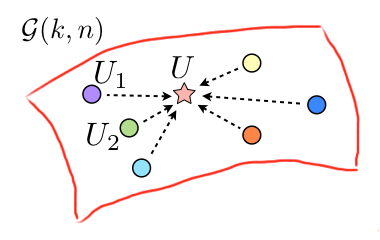}~ \\
			~(a)~ \quad & \quad ~(b)~ \\
		\end{tabular}
	\end{center}
	\caption{Illustrations of graph layer merging. (a) Co-regularization \cite{Kumar11a}: iterative update of the individual subspace representations. The upper index $[N]$ represents the number of iterative steps on each individual subspace representation. The final update of the subspace representation for the most informative graph ($U_1^{[N]}$, shown as a star) is considered as a good combination; (b) Proposed merging framework: the representative subspace ($U$, shown as a star) is found in one step.}
	\label{fig:approaches}
\end{figure*}

\begin{figure*}[t]
	\begin{center}
		\begin{tabular}{cc}
			~\includegraphics[width=0.60\textwidth]{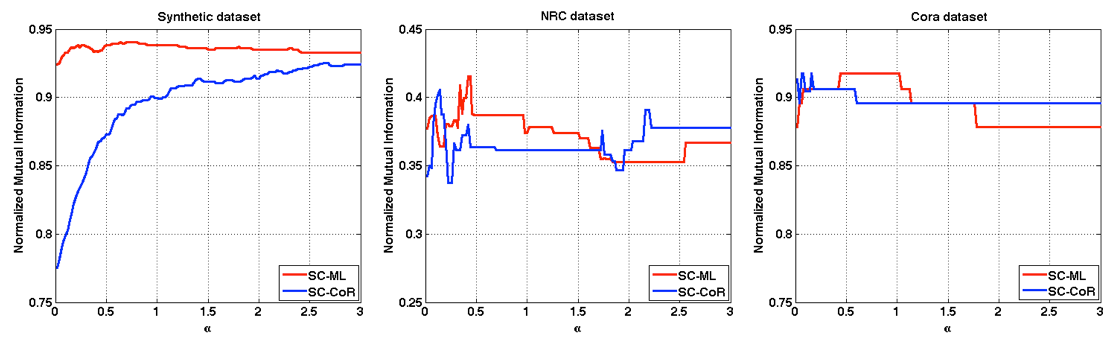}~\\
		\end{tabular}
	\end{center}
	\caption{Performances of \textbf{SC-ML} and \textbf{SC-CoR} under different values of parameter $\alpha$ in the corresponding implementations.}
	\label{fig:parameter}
\end{figure*}

We now take a closer look at the comparisons between \textbf{SC-ML} and \textbf{SC-CoR}. Although the latter is not developed from the viewpoint of subspace analysis on the Grassmann manifold, it can actually be interpreted as a process in which individual subspace representations are updated based on the same distance analysis as in our framework. In this sense, \textbf{SC-CoR} uses the same distance as ours to measure similarities between subspaces. The merging solution however leads to a different optimization problem than that of Eq. (\ref{eqn:merge}), which is based on a slightly different merging philosophy. Specifically, it enforces the information contained in the individual subspace representations to be consistent with each other. An alternating optimization scheme optimizes, at each step, one subspace representation, while fixing the others. This can be interpreted as a process in which one subspace at each step becomes closer to other subspaces in term of the projection distance on the Grassmann manifold. 
Upon convergence, all initial subspaces are ``brought" closer to each other and the final subspace representation from the most informative graph layer is considered as the one that combines information from all the graph layers efficiently. Two illustrations of \textbf{SC-CoR} and \textbf{SC-ML} are shown in Fig.~\ref{fig:approaches} (a) and (b), respectively. Therefore, on the one hand, results for both approaches demonstrate the benefit of using our distance analysis on the Grassmann manifold for merging information in multi-layer graphs. Indeed, for both approaches, since the distances between the solutions and the individual subspaces are minimized without sacrificing too much of the information from individual graph layers, the resulting combinations can be considered as good summarizations of the multiple graph layers. On the other hand, however, \textbf{SC-ML} differs from \textbf{SC-CoR} mainly in the following aspects. First, the alternating optimization scheme in \textbf{SC-CoR} focuses only on optimizing one subspace representation at each step, and it requires a sensible initialization to guarantee that the algorithm ends up at a good local minimum for the optimization problem; it also does not guarantee that all the subspace representations converge to one point on the Grassmann manifold (it uses the final update of the most informative layer for clustering)\footnotemark[8]. In contrast, \textbf{SC-ML} directly finds a single representation through a unique optimization of the representative subspace with respect to all graph layers jointly, which does not need alternating optimization steps and careful initializations. These are the possible reasons that explain why \textbf{SC-ML} performs better than \textbf{SC-CoR} in our experiments, as we can see in Table~\ref{tab:results}. Second, it is worth noting that, from a computational point of view, the optimization process involved in \textbf{SC-ML} is much simpler than that in \textbf{SC-CoR}. Specifically, the iterative nature of \textbf{SC-CoR} requires solving an eigenvalue problem for $MN$ times, where $M$ and $N$ are the number of individual graphs and the number of iterations needed for the algorithm to converge, respectively. In contrast, since \textbf{SC-ML} aims at finding a globally representative subspace without modifying the individual ones, it needs to solve an eigenvalue problem only once.
\footnotetext[8]{In \cite{Kumar11a}, the authors have also proposed a ``centroid-based co-regularization approach" that introduces a consensus representation. However, such a representation is still computed via an alternating optimization scheme, which needs a sensible initialization and keeps the same iterative nature.}
 
Finally, we discuss the influence of the choice of the regularization parameter $\alpha$ on the performance of \textbf{SC-ML}. In Fig.~\ref{fig:parameter}, we compare the performances of \textbf{SC-ML} and \textbf{SC-CoR} in terms of \textit{NMI} under different values of parameter $\alpha$ in the corresponding implementations. As we can see, in our experiments, \textbf{SC-ML} achieves the best performances when $\alpha$ is chosen between 0.4 and 0.6, and it outperforms \textbf{SC-CoR} for a large range of $\alpha$ for the synthetic and NRC datasets. For the Cora dataset, the two algorithms achieve the same performance at different values of $\alpha$, but \textbf{SC-ML} permits a larger range of parameter selection. Furthermore, it is worth noting that the optimal values for $\alpha$ in \textbf{SC-ML} lie in similar ranges across different datasets, thanks to the adoption of the normalized graph Laplacian matrix whose spectral norm is upper bounded by 2. In summary, this shows that the performance of \textbf{SC-ML} is reasonably stable with respect to the parameter selection.

\section{Conclusions}
\label{sec:conclusion}
In this paper, we provide a framework for analyzing information provided by multi-layer graphs and for clustering vertices of graphs in rich datasets. Our generic approach is based on the transformation of information contained in the individual graph layers into subspaces on the Grassmann manifold. The estimation of a representative subspace can then be essentially considered as the problem of finding a good summarization of multiple subspaces using distance analysis on the Grassmann manifold. The proposed approach can be applied to various learning tasks where multiple subspace representations are involved. Under appropriate and realistic assumptions, we show that our framework can be applied to the clustering problem on multi-layer graphs and that it provides an efficient solution that is competitive to the state-of-the-art techniques. Finally, we mention the following research directions as interesting and open problems. First, the subspace representation inspired by spectral clustering is not the only valid representation for the graph information. As suggested by the works in \cite{Newman06a,Newman06b}, the eigenvectors of the modularity matrix of the graph can also be used as low dimensional subspace representation for the information contained in the graph. Therefore, an interesting problem is to find the most appropriate subspace representation for the data available, either they are graphs or of some more general forms. Second, we believe that better clustering performance can be achieved if prior information on the data is available, in particular about the consistency of the information in the different graph layers. These problems are however left for future studies.

\section{Acknowledgement}
This work has been partly supported by Nokia Research Center (NRC) Lausanne, and the EDGAR project funded by Hasler Foundation, Switzerland.

\bibliographystyle{IEEEtran.bst}
\bibliography{mybibfile.bib}

\end{document}